\documentclass[]{spie}

\usepackage{amsmath,amsfonts,amssymb}
\usepackage{graphicx}
\usepackage[utf8]{inputenc}
\usepackage[T1]{fontenc}
\usepackage{hyperref}
\usepackage{url}
\usepackage{booktabs}
\usepackage{nicefrac}
\usepackage{microtype}
\usepackage{xcolor}
\usepackage{subcaption}
\usepackage{csquotes}

\usepackage[disable]{todonotes}

\usepackage{enumitem}
\newlist{itemtight}{itemize}{5}
\setlist[itemtight]{label=\textbullet, itemsep=0pt}
\newlist{enumtight}{enumerate}{5}
\setlist[enumtight]{label*=\arabic*., itemsep=0pt}

\hypersetup{		
			colorlinks=true,
			linkcolor=red,
            linktoc=page,
			anchorcolor=black,
			citecolor=gray,
			urlcolor=blue,
			pdftitle={Learning Local Causal World Models with State Space Models and Attention},
			pdfauthor={Francesco Petri}
}

\newcommand{\R}{\mathbb{R}}
\newcommand{\acronym}{S2-SSM}
\newcommand{\fullname}{Sparse Slot State Space Model}

\title{Learning Local Causal World Models with State Space Models and Attention}

\author[a,b,*]{Francesco Petri}
\author[a,c]{Luigi Asprino}
\author[a,c]{Aldo Gangemi}
\affil[a]{Institute for Cognitive Sciences and Technologies, National Research Council, Rome, Italy}
\affil[b]{Sapienza University of Rome, Italy}
\affil[c]{University of Bologna, Italy}

\authorinfo{
*Corresponding author: \url{francesco.petri@uniroma1.it}
}

\begin{document} 
\maketitle

\begin{abstract}
World modelling, i.e. building a representation of the rules that govern the world so as to predict its evolution, is an essential ability for any agent interacting with the physical world. Despite their impressive performance, many solutions fail to learn a causal representation of the environment they are trying to model, which would be necessary to gain a deep enough understanding of the world to perform complex tasks. With this work, we aim to broaden the research in the intersection of causality theory and neural world modelling by assessing the potential for causal discovery of the State Space Model (SSM) architecture, which has been shown to have several advantages over the widespread Transformer. We show empirically that, compared to an equivalent Transformer, a SSM can model the dynamics of a simple environment and learn a causal model at the same time with equivalent or better performance, thus paving the way for further experiments that lean into the strength of SSMs and further enhance them with causal awareness.
\end{abstract}

\keywords{World modelling, causal discovery, state-space models, object-oriented representations}

\section{INTRODUCTION}
\label{sec:intro}

World modelling is the ability of an artificial agent to build an internal representation of the world in which it operates.
This representation is employed by the agent to forecast  the evolution of the world.
The problem of building a world model spans many branches of artificial intelligence, such as planning and reinforcement learning \cite{micheli23}, physics modelling and reasoning \cite{physics-causal:li2022aaai}, and robotics \cite{wm-in-robotics:wu2022corl}.
An accurate representation of the world allows building simulations that, in turn, enable practitioners to gather additional data and test the performance of an artificial agent without interacting with their environment.
This is convenient because interacting with an agent's environment can be time-consuming, risky due to possible failures of physical components, and sometimes even impossible due to the potential unavailability of the environment (e.g. experimenting with the exploration of Mars by a rover).

Recent efforts have produced highly accurate and efficient world models in a variety of environments \cite{micheli23, slotformer:wu23}. 
In particular, the State-Space Model (SSM) is a new type of neural architecture that has the ability to remember past observations over extremely long time horizons, allowing it to model large environments and long-range interactions \cite{s4:deng2023nips}. 
In addition, it has recently been shown that learning a causal model of the environment, i.e. a model that describes the causal relationships between objects in the environment, is necessary to achieve a good level of robustness to changes in the world \cite{motivation:richens2024iclr}.
In this paper, we investigate whether SSM can learn causal models and the benefits of doing so in the world modelling task.

Following the SSM architectural style, we propose a novel model called \fullname{} (\acronym{}) which features a sparsity regularization technique originally designed for the Transformer architecture \cite{spartan:lei2024arxiv} to learn a local causal model of the interactions between the objects at each moment in time.
We experiment with the world modelling and causal discovery tasks in a synthetic environment.
We test the performance of \acronym{} against a Transformer-based architecture, i.e. SPARTAN~\cite{spartan:lei2024arxiv}.
We report preliminary results showing that \acronym{} is able to learn an accurate causal model. 
This contributes to enhancing the world modelling capability of the model and surpassing  the performance of the Transformer architecture, or in the worst case scenario, matching  it.

\section{Related work}
\label{sec:related}

\subsection{Causal Discovery}

Integrating a general theory of causation into machine learning methods is an ongoing research effort.
Learning a causally informed representation of the data would enable a deeper understanding of the interactions between the entities described in the data, whether the domain is a physical world populated with objects \cite{physics-causal:li2022aaai} or something more abstract like language \cite{DBLP:conf/coling/WangZSYGZH22}, which would increase the interpretability of the resulting model.

Several formal structures have been developed to study causal relationships between variables. 
Examples are Bayesian networks, which are widely used method for causal learning through statistical analysis \cite{bayesian-nets-heckerman}, and Structural Causal Models, which are expressed as a graph where each node represents a causal variable and the edges indicate cause-effect relations \cite{scm-pearl}. 
In particular, graph-based models are frequently combined with the attention operation in machine learning.
The attention operation determines the relative importance (called weight) of each component in a sequence relative to the other components in that sequence. 
The attention weights function as a metric for the degree of correlation between two variables, thereby indicating the probability of a causal relationship between them within the graph model~\cite{yu2023ijcai, DBLP:conf/nips/PitisCG20}. 
Transformers is a neural architecture, which has been applied in multiple domains, that relies on a multi-head attention mechanism~\cite{transformer:vaswani17}.
In particular, Transformers have been utilised in the context of learning graph causal models~\cite{spartan:lei2024arxiv}. 
In this study, we further explore the use of attention weights as indicators of causal relationships between variables in a different architectural framework, namely the state space models, which is particularly relevant to the task of world modelling. 

\subsection{World Modelling}

The world modelling problem has received tremendous attention in the last few years. In reinforcement learning, being able to simulate the dynamics of the environment is especially useful, because it enables the agent to act and learn in its own simulated world without paying the cost of interacting with the \enquote{real} environment. The Dreamer algorithm
\cite{dreamerv2:hafner2021iclr} has been relatively influential on the topic, including applications in robotic domains \cite{wm-in-robotics:wu2022corl}, where real-world interaction has the unfortunate potential of breaking expensive equipment, in addition to time costs.
Additional approaches in reinforcement learning include solutions based on causal discovery and reasoning \cite{yu2023ijcai}, which aim at learning a causal model of the environment to better understand the interactions between the agent and the world, and even provide an explanation for the actions taken by the agent.

Many Transformer-based approaches have been studied, due to their generally good performance in different tasks and the sample efficiency they provide in this specific problem \cite{micheli23, slotformer:wu23, spartan:lei2024arxiv}.
Recently, the newer State-Space Model architecture (SSM) has also been applied to this problem \cite{s4:deng2023nips}. This sequence model based on principles from control theory has already been shown to have some advantages over Transformers, in particular a high inference efficiency and a great memory capacity, and they achieve similar levels of performance while not suffering from the quadratic memory cost that notoriously afflicts Transformers \cite{mamba:gu2024colm, s4:deng2023nips}.
We particularly note the very recent SlotSSM \cite{slotssm:jiang2024neurips}, which exploits object-oriented representations and attention to model the interaction between objects in a scene. Despite this interest, the adeptness of SSMs for causal learning has not been extensively assessed yet. This work aims to contribute in this direction.

\section{METHOD}
\label{sec:method}

We address the problem of learning causal world models of an environment for which a video of its objects is available. 
To do so we empower a SlotSSM-based world model \cite{slotssm:jiang2024neurips} with the ability to capture local causation relations between objects in a scene. 
We call the resulting architecture \textbf{\acronym{}} (\fullname{}).

At its core, the problem of world modelling for video can be expressed as the task of predicting the next frame of a video, i.e. $x_{t+1} \in \R^{W \times H \times C}$, given one or more previous frames $x_1, ..., x_t$. 
Indeed, in order to predict the subsequent scene in a video, it is necessary to comprehend the underlying principles of the observed environment \cite{dreamerv2:hafner2021iclr}.

Rather than modelling low-level signal of the images (i.e. pixels) directly, we first extract a high-level representation based on the objects in the scene. 
Given the t-th frame of the sequence $x_t$, we extract a representation $s_t$ of the $O$ objects of the scene. 
The representation $s_t$ is the concatenation of $O$ vectors (one for each object of the scene, i.e. $s_t = s_t^1, ..., s_t^O$ where $s_t^i \in \R^D$ with $D$ is the dimension of the latent space.
The latent representation of objects $s_t^i$  is called \textit{slot}  \cite{slot-attn:locatello2020nips, slotformer:wu23}.

\begin{figure}[t]
\includegraphics[width=\textwidth]{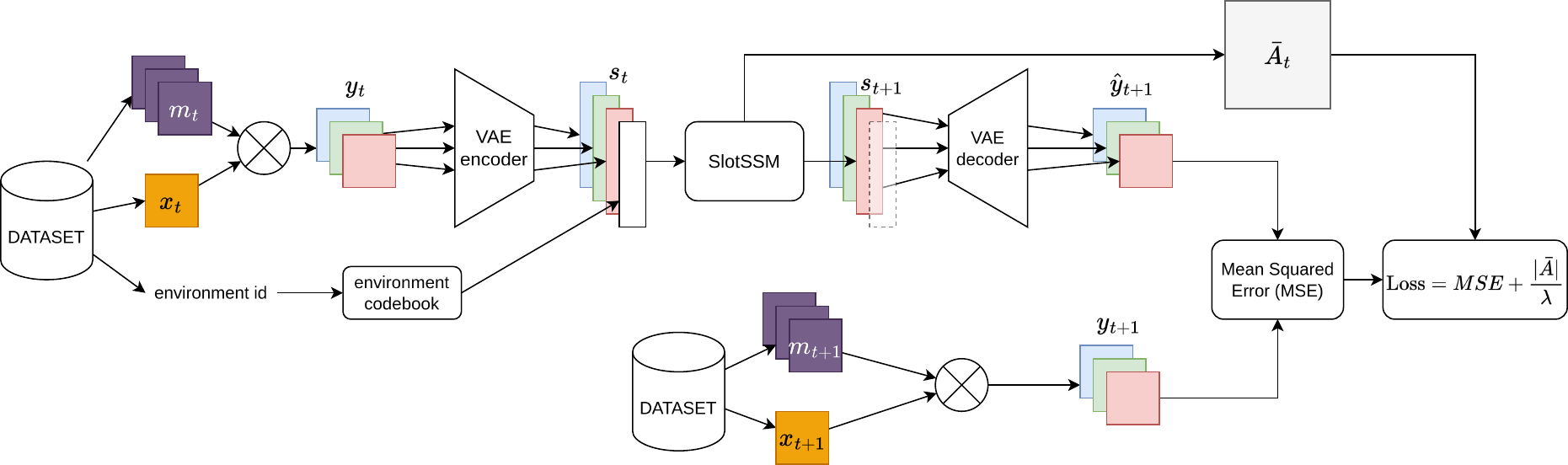}
\caption{Scheme of the \acronym{} architecture.}
\label{fig:architecture}
\end{figure}

\subsection{Obtaining Object Representations}

As the learning of object-oriented representations is not the primary focus of this work, the procedure is simplified by the provision of masks for each object with the dataset.
These masks are then used to isolate each object in the scene. 
Finally, a variational autoencoder (VAE) is employed to encode each object into a $D$-dimensional slot:
\[
y_t^{1,...,O} = x_t \cdot m_t^{1,..,O}; \qquad
s_t^{1,...,O} = \text{Encoder}(y_t^{1,...,O}); \qquad
\hat y_t^{1,...,O} = \text{Decoder}(s_t^{1,...,O})
\]
The VAE is trained at the same time as the world model in an end-to-end fashion.

\subsection{Environment Representation}

In addition to the object slots, the world model is also provided a representation of the environment, which also takes the form of a vector $s^{env} \in \R^D$. 
Since this is a representation of the scene along with the objects, we call this vector the \textit{environment slot} while we refer to the object representation as \textit{object slots}.
In general, there may be multiple environments each of which is subject to a different set of rules.

During the training phase, each environment slots are provided by a learnable codebook, so that the model is trained to detect different environments without prior information of what specific rules that govern them and how the objects are affected. 
During the testing phase,  the model is not provided with any information regarding the environment. 
The model identifies the current environment as the one that best fits the data.
Nevertheless, in the event of the environment being encountered for the first time (i.e. if it was not included in the training dataset), the model has the capacity to learn a new representation of it during the training phase.

\subsection{World Model}

This module takes as input the concatenation of the object and environment slots at time t $(s_t, s^{env})$ and predicts the future state $s_{t+1}$. 
It is composed of a number of SlotSSM blocks, each featuring an SSM layer (Mamba) \cite{mamba:gu2024colm} to predict the free evolution of each object (i.e. independently of the others) followed by a cross-attention layer to model the interaction between the slots. The sparsity regularization (see \autoref{sub:regularization}) is applied to each of such attention layers.

In the evaluation against Transformer baseline, we replace this module with a Transformer encoder, in accordance with the methodology outlined in previous work \cite{spartan:lei2024arxiv}.
The remainder of the architecture remains unaltered. 
In this case, the regularization is applied to the attention layers of the Transformer.

\subsection{Learning a Causal Graph}

We assume that the object and environment slots represent the causal variables governing the evolution of the environment's state. For example, the interaction of two objects might alter a slot, or a special rule of the current environment may cause an entity to behave differently then it normally would (see \autoref{sec:experiments} for a more concrete description based on the world used in our experiments).

Following existing literature \cite{spartan:lei2024arxiv, yu2023ijcai, DBLP:conf/nips/PitisCG20}, we use the attention weights between the slots, which in our case are calculated by the SlotSSM interaction layers, as a measure of the causal interaction between the variables.

Each single attention layer takes as input the slots $(s^1, ..., s^O, s^{env})$, extracts the query, key, and value vectors $\{q_i, k_i, v_i\}$ for each via linear projection, and construct the adjacency matrix for the causal graph based on the cosine similarity between the queries and keys:
\[
A_{ij} = \sigma(q_i^T k_j)
\]
Then, such adjacency matrix is used as a mask for the attention operation, so that the slots at time $t+1$ are determined by the pairs of variables that have a strong causal interaction:
\[
s_{t+1}^i = s_t^i + \sum_j A_{ij} \frac{exp(q_i^T k_j) / \sqrt{D}}{\sum_k exp(q_k^T k_j) / \sqrt{D}} v_j \quad \text{for } i, j = 1, ..., O, env
\]
In the case of $L > 1$ attention layers, we have to consider indirect connections across multiple layers \cite{spartan:lei2024arxiv}. To that end, the final causal representation output by the world model is the number of \textit{paths} between each pair of variables across the adjacency matrix of all layers:
\[
\bar A = (A^L + \mathbb{I})...(A^2 + \mathbb{I})(A^1 + \mathbb{I})
\]
where $A^i$ is the adjacency matrix for the i-th layer and  $\mathbb{I}$ is the identity matrix, representing the self-loop that each variable has by default, due to the residual connection at the end of the attention operation.

The learned causal model can be assimilated to a structural causal model \cite{scm-pearl} where the object and environment slots act as causal variables, the graph is described by the $A$ matrix, and the transition functions for each variable are learned and approximated by the world model.

\subsection{Training Objective}
\label{sub:training-objective}

During training, the whole \acronym{} architecture aims to minimize a loss that reflects our objective to learn a world model that can both predict the next frame in the video and represent the transition with a minimal causal model. As such, the loss function has two components, regulated by a balancing parameter: a reconstruction loss and a regularization on the causal graph.
\[
\mathcal{L} = MSE(\hat y_{t+1}^{1,...,O}, y_{t+1}^{1,...,O}) + \frac{|\bar A|}{\lambda}
\]

\subsubsection{Reconstruction loss}

The world modelling objective is trained via the reconstruction of the next frame in pixel space. Specifically, to ensure accurate modelling of each object, the reconstruction loss is calculated on isolated images of each object, rather than their combination (see the end of \autoref{fig:architecture}).

We use the Mean Squared Error (MSE) as the error function.

\subsubsection{Sparsity regularization}
\label{sub:regularization}

To ensure that the model learns the minimal local causal model that explains the provided observations of the environment, thus avoiding superfluous or erroneous connections, we apply a penalty on the training objective that scales with the total number of paths in the causal graphs of each attention layer, as calculated by the $\bar A$ matrix.

The relative importance of this term compared to the reconstruction error is represented by the $\lambda$ parameter. This plays a vital role to make sure both components of the loss are attended to while training: if the regularization has too little weight, the model will end up either not learning a causal model at all, but if it is made too important, the model will be discouraged from making any connections in the attention layer and will not be able to function as a world model.

Based on previous work \cite{spartan:lei2024arxiv}, we adopt a dynamic $\lambda$ that starts very high and changes over time depending on the reconstruction error:
\[
\lambda \xleftarrow{} e^{(MSE - \tau)} \cdot \lambda
\]
Where the target error $\tau$ is a constant determined as the reconstruction error achieved by a simpler, non-causal model, such as a fully connected neural network.

This acts like a schedule driven by the reconstruction error: initially, the training focuses on the world modelling objective, then once the reconstruction error reaches a baseline value, the regularization on the causal graph gains more importance, thus driving the model to learn an essential causal model while further improving the reconstruction.

\section{EXPERIMENTS}
\label{sec:experiments}

\begin{figure}[t]
\begin{subfigure}[t]{0.49\textwidth}
    \centering
    \includegraphics[width=0.5\textwidth]{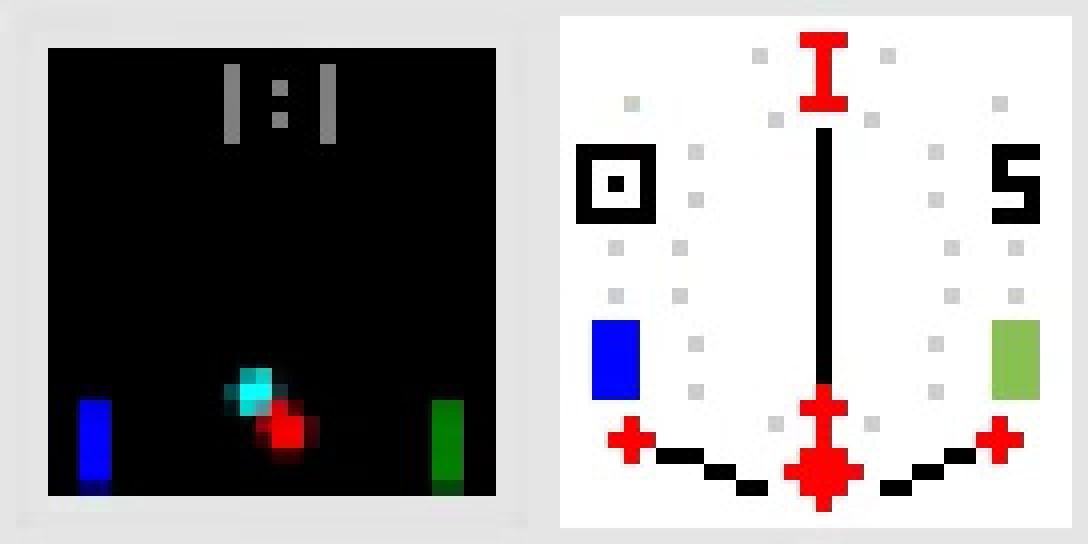}
    \caption{Env: ball slows in the middle.}
    \label{fig:q1}
\end{subfigure}
\hfill
\begin{subfigure}[t]{0.49\textwidth}
    \centering
    \includegraphics[width=0.5\textwidth]{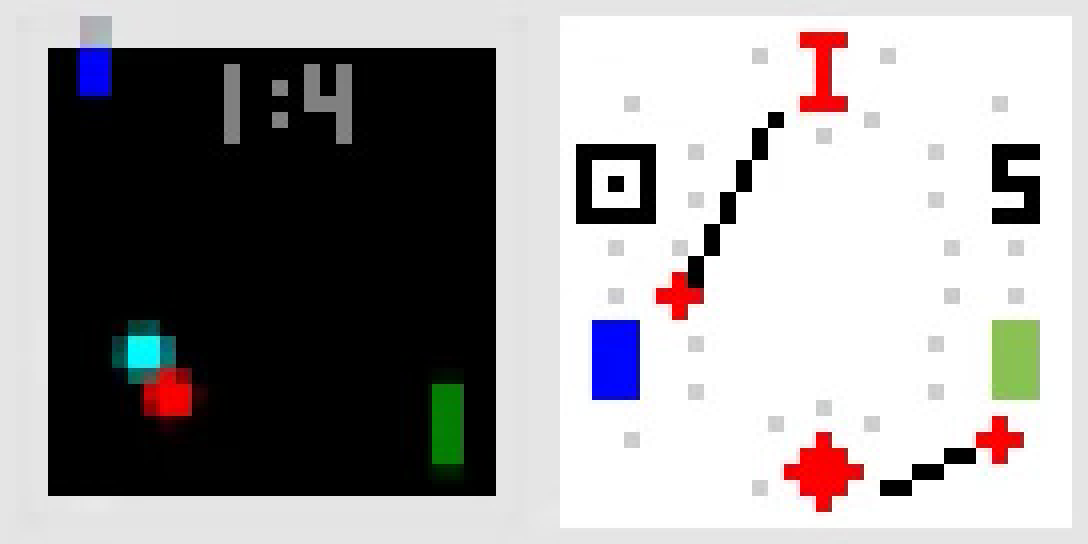}
    \caption{Env: left pong moves away from the ball.}
    \label{fig:q2}
\end{subfigure}
\caption{Qualitative examples showing the reconstructed image and the causal graph in two different environments. The nodes of the graph, from the top clockwise, are: environment slot, score, right pong, ball, left pong, border.}
\label{fig:qual}
\end{figure}

We assess the performance of the \acronym{} model qualitatively through examining the reconstructed videos and the causal graph produced by the model and quantitatively through the following experiments:

\begin{itemtight}
    \item Comparison of \acronym{} with a Transformer architecture based on the same principles of causal discovery and sparsity.
    \item An ablation study where we remove the sparsity regularization and show that it is essential for learning a correct causal graph.
\end{itemtight}

Much of the experimental setup was intentionally designed to comply with the reference paper \cite{spartan:lei2024arxiv}.

\subsection{Experimental Setup}

\subsubsection{Environment description}

We train and evaluate the model on 32x32 videos from the Interventional Pong dataset \cite{citris:lippe22a}, a simple synthetic world whose underlying physics are easy to manipulate (\autoref{fig:qual}). The objects of interest are the following:

\begin{itemtight}
    \item The red ball moves at a constant speed and bounces off the border and the pongs. A cyan circle is also added to the images to represent the velocity of the ball.
    \item The left and right pongs, colored in blue and green respectively, are located on their respective sides of the screen and move up and down to follow the ball (up to a maximum velocity). This is interpreted as the ball having a causal influence on the pongs' behavior.
    \item The score counter does not interact with anything, but it is updated whenever the ball crosses the left or right sides of the screen, thus scoring a point.
    \item The border is completely inert.
\end{itemtight}

We consider 11 different environments: 7 \textit{simple} ones, including the default environment where there are no changes, and 4 \textit{composite} ones. Following the previous work on causal discovery with this dataset, \cite{spartan:lei2024arxiv}
each (non-default) environment features changes that represent altered physics or behaviors in the world.
The model is trained only on the simple environments and tested on each of them, simple and composite alike, separately, so that the composite environments represent unknown situations to evaluate the adaptability of the model.

The full list of environments is reported in \autoref{tab:interventions}.

\subsubsection{Experimental procedure}

One test consists of two steps. First, we select one intervention (unknown to the model) and reserve a small portion of its dataset (10\%) for a few-shot fine-tuning phase aimed at learning an intervention slot that best represents the observations. This lasts only for a few hundred training steps, and only the intervention slot is learned: the pre-trained VAE and world model remain frozen. The learning objective here is based exclusively on the reconstruction error, i.e. like in \autoref{sub:training-objective} with $\lambda = \infty$.

Next, we evaluate the performance of the model using the rest of the data for the selected intervention and the intervention slot just learned in the fine-tuning phase. Qualitatively, we are interested in observing the quality of the reconstructed image and a causal graph that corresponds to the interactions happening throughout the video. Quantitatively, we base our evaluation on two metrics:

\begin{itemtight}
    \item Mean squared error (MSE) on the reconstruction of the next frame. Just like in the training phase, this error is calculated on each single-object masked image.
    \item Structural Hamming distance (SHD) \cite{shd:Tsamardinos2006} on the causal graph to evaluate its correctness. It is worth noting that this is \textit{not} an explicit component of the training objective: the world model is only encouraged to find the smallest causal graph that enables an accurate reconstruction of the input.
\end{itemtight}

We repeat this test for each of the interventions we consider in this work, and report our results separately by intervention as well as an average of all the experiments.

\begin{table}[b]
\centering
\begin{tabular}{lcc}
    \hline
     & Mean Squared Error & Structural Hamming Distance \\
    \hline
    S2-SSM (Ours) & $\mathbf{2.90 \cdot 10^{-4}}$ & $\mathbf{1.41}$ \\
    S2-TE & $5.27 \cdot 10^{-4}$ & $11.15$ \\
    dense-SSM & $3.46 \cdot 10^{-4}$ & $16.40$ \\
    dense-TE & $5.34 \cdot 10^{-4}$ & $16.81$ \\
    \hline
\end{tabular}
\caption{Average metrics for our experiments. This is the same data as the MEAN columns in \autoref{fig:results}.}
\label{tab:results}
\end{table}

\subsection{Results}

\begin{figure}[t]
\begin{subfigure}[t]{0.49\textwidth}
    \includegraphics[width=\textwidth]{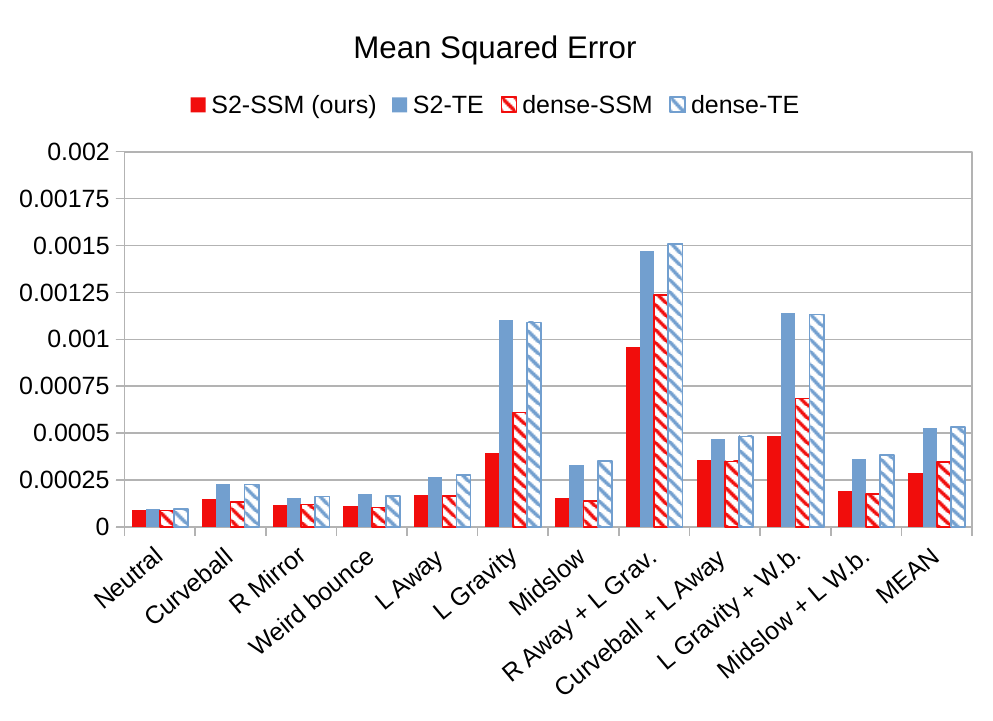}
    \caption{Mean Squared Error (MSE).}
    \label{fig:mse}
\end{subfigure}
\hfill
\begin{subfigure}[t]{0.49\textwidth}
    \includegraphics[width=\textwidth]{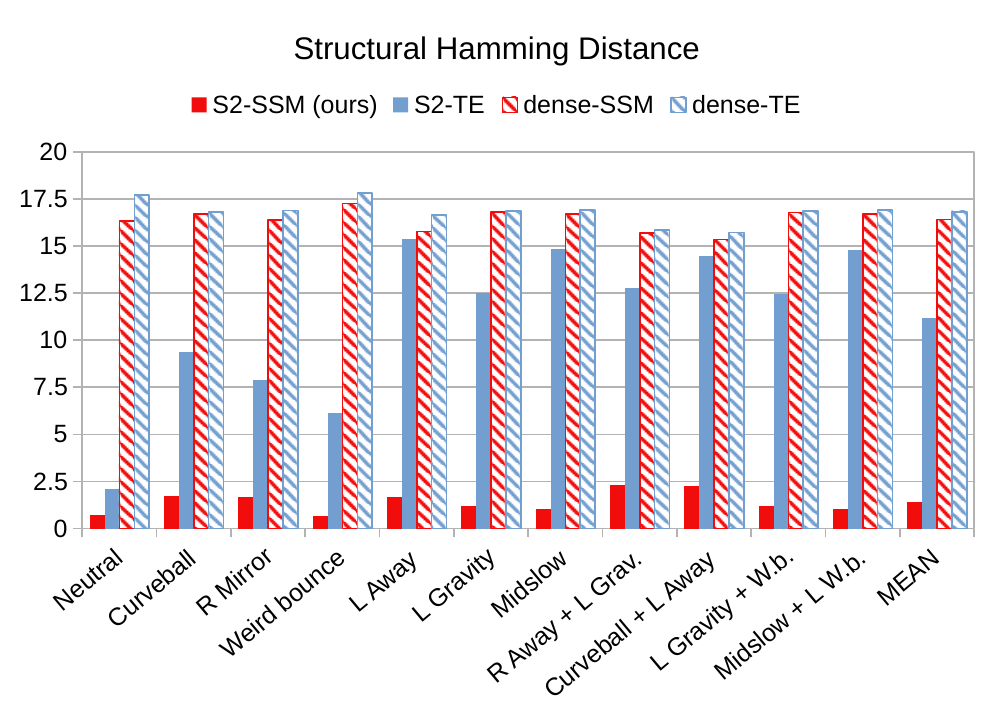}
    \caption{Structural Hamming Distance (SHD).}
    \label{fig:shd}
\end{subfigure}
\caption{Error metrics separated by environment. The MEAN column is the average of all the others.}
\label{fig:results}
\end{figure}

We report the results of the experiments described above in \autoref{fig:results} and \autoref{tab:results}, where \acronym{} is our proposed SSM-based architecture with sparsity regularization, S2-TE (Sparse Slot Transformer Encoder) is the Transformer-based baseline, and dense-SSM and dense-TE are the respective variants without sparsity regularization for the purposes of the ablation study.

\subsubsection{Comparison between SlotSSM and Transformer}

Focusing on the comparison between \acronym{} and S2-TE at first, we notice that, in regard to both metrics, the SSM's performance is either competitive with or better than the Transformer's (depending on the environments), showing that the SlotSSM architecture is capable of learning a causal understanding of the world it models more effectively than its Transformer counterpart.

It should be noted that, although an effort was made to create an experimental setup that is as close as possible to the work of causal discovery with sparsity-regulated Transformers by Lei et al. \cite{spartan:lei2024arxiv}, the source code for those experiments was not released,
so the results we observe for the S2-TE model differ from the ones reported in that work. This is likely due to differences in the implementations of the architecture and the experiments. Still, we believe that the results of our experiments support our claim, and the average SHD we observe for \acronym{} (\autoref{tab:results}) is very close to that reported by Lei et al. (1.51 as per their table 1).

\subsubsection{The importance of sparsity regularization}

As for the ablation study, we compare \acronym{} with dense-SSM and S2-TE with dense-TE. Regardless of the underlying world model architecture, we observe that removing the sparsity regularization has a slightly negative effect on the reconstruction metric (MSE) and drastically worsens the quality of the causal graph (SHD) to the point of losing any meaning, as the model just predicts a causal relation between all pairs of objects in the scene.

This shows that, although the world modeling performance does not seem to be greatly affected, regularizing the causal graph during training is crucial to learning a meaningful causal understanding of the environment.

\begin{table}[b]
\centering
\begin{tabular}{ll|l}
    \hline
    Simple env.s & Description & Composite env.s \\
    \hline
    Neutral & No alterations & \\
    Curveball & Curved ball trajectory & Curveball + L Away \\
    L Away & Left pong moves away from ball & R Away + L Gravity\\
    R Mirror & Right pong position is mirrored & L Gravity + W.b. \\
    Midslow & Ball slows down in the middle section & Midslow + L W.b. \\
    Weird bounce (W.b.) & Ball turns 180° when bouncing off a pong & \\
    L Gravity & Ball has a gravity acceleration to the left & \\
    \hline
\end{tabular}
\caption{The different environments used in to train and experiment with the model.}
\label{tab:interventions}
\end{table}

\section{DISCUSSION}
\label{sec:discussion}

In this work, we detailed the \fullname{} world model (\acronym{}), a SSM-based architecture that learns a causal world model through a sparsity regularization on its interaction layers. We performed experiments on a synthetic environment and showed that it is capable of accurately predicting the future state of the environment and providing an equally accurate graphical causal model of the interactions between the objects in the environment. The tests on composite environments also show the ability of \acronym{} to adapt to previously unseen environments. The performance of the our model is better, in the worst case comparable to a previous work applying the same technique to a Transformer-based architecture \cite{spartan:lei2024arxiv}, showing the viability of SSM-based world models for causal discovery and learning.

\subsection{Future Work}

Having established that it is possible for SSM-based world models to capture causal relations in the same way as Transformer-based ones can, we now look to demonstrating that the former can also leverage their typical strengths without losing such properties. For example, since SSMs have been found to have a high memory capacity, \cite{s4:deng2023nips}
we plan to experiment with environments where the behaviour of the entities and the interventions of the causal relationships are determined by previous observations, rather than simply being given as an intervention slot. In a similar vein, we are also interested in studying whether these architectures can model objects that have been observed in the past but are out of view at present, e.g. by adding an occlusion feature to the scenes.

\acknowledgments   
 
This work has been funded by the Italian National PhD Program on Artificial Intelligence run by Sapienza University of Rome in collaboration with the Italian National Council for Research, and by the Italian PNRR MUR project PE0000013-FAIR.

\bibliographystyle{spiebib}
\bibliography{0_main.bbl}

\begin{thebibliography}{10}

\bibitem{micheli23}
Micheli, V., Alonso, E., and Fleuret, F., ``Transformers are sample-efficient world models,'' in [{\em The Eleventh International Conference on Learning Representations, {ICLR} 2023, Kigali, Rwanda, May 1-5, 2023}{\nolinebreak\hspace{0.1em}]},  OpenReview.net (2023).

\bibitem{physics-causal:li2022aaai}
Li, Z., Zhu, X., Lei, Z., and Zhang, Z., ``Deconfounding physical dynamics with global causal relation and confounder transmission for counterfactual prediction,'' in [{\em Thirty-Sixth {AAAI} Conference on Artificial Intelligence, {AAAI} 2022, Virtual Event, February 22 - March 1, 2022}{\nolinebreak\hspace{0.1em}]},   1536--1545, {AAAI} Press (2022).

\bibitem{wm-in-robotics:wu2022corl}
Wu, P., Escontrela, A., Hafner, D., Abbeel, P., and Goldberg, K., ``Daydreamer: World models for physical robot learning,'' in [{\em Conference on Robot Learning, CoRL 2022, 14-18 December 2022, Auckland, New Zealand}{\nolinebreak\hspace{0.1em}]},  {\em Proceedings of Machine Learning Research} {\bf 205},  2226--2240, {PMLR} (2022).

\bibitem{slotformer:wu23}
Wu, Z., Dvornik, N., Greff, K., Kipf, T., and Garg, A., ``Slotformer: Unsupervised visual dynamics simulation with object-centric models,'' in [{\em The Eleventh International Conference on Learning Representations, {ICLR} 2023, Kigali, Rwanda, May 1-5, 2023}{\nolinebreak\hspace{0.1em}]},  OpenReview.net (2023).

\bibitem{s4:deng2023nips}
Deng, F., Park, J., and Ahn, S., ``Facing off world model backbones: Rnns, transformers, and {S4},'' in [{\em Advances in Neural Information Processing Systems 36: Annual Conference on Neural Information Processing Systems 2023, NeurIPS 2023, New Orleans, LA, USA, December 10 - 16, 2023}{\nolinebreak\hspace{0.1em}]},  (2023).

\bibitem{motivation:richens2024iclr}
Richens, J. and Everitt, T., ``Robust agents learn causal world models,'' in [{\em The 12th International Conference on Learning Representations, {ICLR} 2024, Vienna, Austria, May 7-11, 2024}{\nolinebreak\hspace{0.1em}]},  OpenReview.net (2024).

\bibitem{spartan:lei2024arxiv}
Lei, A., Sch{\"{o}}lkopf, B., and Posner, I., ``{SPARTAN:} {A} sparse transformer learning local causation,'' {\em CoRR}~{\bf abs/2411.06890} (2024).

\bibitem{DBLP:conf/coling/WangZSYGZH22}
Wang, S., Zhou, J., Sun, C., Ye, J., Gui, T., Zhang, Q., and Huang, X., ``Causal intervention improves implicit sentiment analysis,'' in [{\em Proceedings of the 29th International Conference on Computational Linguistics, {COLING} 2022, Gyeongju, Republic of Korea, October 12-17, 2022}{\nolinebreak\hspace{0.1em}]},   6966--6977 (2022).

\bibitem{bayesian-nets-heckerman}
Heckerman, D., ``A tutorial on learning with bayesian networks,'' {\em CoRR}~{\bf abs/2002.00269} (2020).

\bibitem{scm-pearl}
Pearl, J., Glymour, M., and Jewell, N.~P.,  [{\em Causal inference in statistics: a primer}{\nolinebreak\hspace{0.1em}]}, Wiley, Chichester, West Sussex (2016).

\bibitem{yu2023ijcai}
Yu, Z., Ruan, J., and Xing, D., ``Explainable reinforcement learning via a causal world model,'' in [{\em Proceedings of the Thirty-Second International Joint Conference on Artificial Intelligence, {IJCAI} 2023, 19th-25th August 2023, Macao, SAR, China}{\nolinebreak\hspace{0.1em}]},   4540--4548, ijcai.org (2023).

\bibitem{DBLP:conf/nips/PitisCG20}
Pitis, S., Creager, E., and Garg, A., ``Counterfactual data augmentation using locally factored dynamics,'' in [{\em Advances in Neural Information Processing Systems 33: Annual Conference on Neural Information Processing Systems 2020, NeurIPS 2020, December 6-12, 2020, virtual}{\nolinebreak\hspace{0.1em}]},  (2020).

\bibitem{transformer:vaswani17}
Vaswani, A., Shazeer, N., Parmar, N., Uszkoreit, J., Jones, L., Gomez, A.~N., Kaiser, L., and Polosukhin, I., ``Attention is all you need,'' in [{\em Advances in Neural Information Processing Systems 30: Annual Conference on Neural Information Processing Systems 2017, December 4-9, 2017, Long Beach, CA, {USA}}{\nolinebreak\hspace{0.1em}]},  (2017).

\bibitem{dreamerv2:hafner2021iclr}
Hafner, D., Lillicrap, T.~P., Norouzi, M., and Ba, J., ``Mastering atari with discrete world models,'' in [{\em 9th International Conference on Learning Representations, {ICLR} 2021, Virtual Event, Austria, May 3-7, 2021}{\nolinebreak\hspace{0.1em}]},  OpenReview.net (2021).

\bibitem{mamba:gu2024colm}
Gu, A. and Dao, T., ``Mamba: Linear-time sequence modeling with selective state spaces,'' in [{\em First Conference on Language Modeling}{\nolinebreak\hspace{0.1em}]},  (2024).

\bibitem{slotssm:jiang2024neurips}
Jiang, J., Deng, F., Singh, G., Lee, M., and Ahn, S., ``Slot state space models,'' in [{\em Advances in Neural Information Processing Systems 38: Annual Conference on Neural Information Processing Systems 2024, NeurIPS 2024, Vancouver, BC, Canada, December 10 - 15, 2024}{\nolinebreak\hspace{0.1em}]},  (2024).

\bibitem{slot-attn:locatello2020nips}
Locatello, F., Weissenborn, D., Unterthiner, T., Mahendran, A., Heigold, G., Uszkoreit, J., Dosovitskiy, A., and Kipf, T., ``Object-centric learning with slot attention,'' in [{\em Advances in Neural Information Processing Systems 33: Annual Conference on Neural Information Processing Systems 2020, NeurIPS 2020, December 6-12, 2020, virtual}{\nolinebreak\hspace{0.1em}]},  Larochelle, H., Ranzato, M., Hadsell, R., Balcan, M., and Lin, H., eds. (2020).

\bibitem{citris:lippe22a}
Lippe, P., Magliacane, S., L{\"o}we, S., Asano, Y.~M., Cohen, T., and Gavves, S., ``{CITRIS}: Causal identifiability from temporal intervened sequences,'' in [{\em Proceedings of the 39th International Conference on Machine Learning}{\nolinebreak\hspace{0.1em}]},  {\em Proceedings of Machine Learning Research} {\bf 162},  13557--13603, PMLR (17--23 Jul 2022).

\bibitem{shd:Tsamardinos2006}
Tsamardinos, I., Brown, L.~E., and Aliferis, C.~F., ``The max-min hill-climbing bayesian network structure learning algorithm,'' {\em Mach. Learn.}~{\bf 65}(1),  31--78 (2006).

\end{thebibliography}

\end{document}